%% file: 00_asmeconf-authorgrid-example.tex
\DocumentMetadata{%
	pdfstandard=A-3u,
	pdfversion=1.7,
	lang=en-US,
}

\documentclass[grid,balance,colorlinks,upint,subscriptcorrection,varvw,hyphenate,vietnamese,greek,russian,german,spanish]{asmeconf}

\hypersetup{%
	pdfauthor={Yifei Li},									  
	pdftitle={ASME Conference Paper LaTeX Template},                  
	pdfkeywords={ASME conference paper, LaTeX template, BibTeX style},
	pdfsubject = {Describes the asmeconf LaTeX template},			  
}

\allowdisplaybreaks 

\usepackage{hologo} 
\usepackage{graphicx}
\usepackage{amsmath}
\usepackage{mathtools,bbm}
\usepackage{array,tabularx}
\usepackage{multirow}
\usepackage{soul}
\usepackage[normalem]{ulem}
\usepackage{boldline}
\usepackage{algorithm,algorithmic}
\usepackage{enumitem}
\usepackage{adjustbox}
\usepackage{booktabs}
\usepackage{listings}
\usepackage{cleveref}
\usepackage{url}
\usepackage{bm}
\usepackage{mathrsfs}
\usepackage{tabularx}

\begin{document}


\ConfName{Proceedings of the ASME 2026 21st International\linebreak Manufacturing Science and Engineering Conference}
\ConfAcronym{MSEC2026}
\ConfDate{June 14--18, 2026} 
\ConfCity{State College, Pennsylvania}
\PaperNo{MSEC2026-182990}

%

\title{Intelligent Navigation and Obstacle-Aware Fabrication for Mobile Additive Manufacturing Systems}

 
%
%
%

\SetAuthors{%
	Yifei Li\affil{1}\CorrespondingAuthor{}, 
	Ruizhe Fu\affil{1}, 
    Huihang Liu\affil{1}, 
	Guha Manogharan\affil{1}, 
	Feng Ju\affil{3}, 
	Ilya Kovalenko\affil{1}
}

\SetAffiliation{1}{The Pennsylvania State University, University Park, PA}
\SetAffiliation{2}{Department of Electrical Engineering and Computer Science\\ The Pennsylvania State University, University Park, PA}
\SetAffiliation{3}{ Arizona State University, Tempe, AZ}
\SetAffiliation{4}{Department of Mechanical Engineering and the Department of Industrial and Manufacturing Engineering\\ The Pennsylvania State University, University Park, PA}

\maketitle

\versionfootnote{Documentation for \texttt{asmeconf.cls}: Version~\versionno, \today.}



\begin{abstract}
\input{01_abstract}

\end{abstract}

\section{Introduction}
\label{sec:intro}
\input{02_intro}

\section{Mobile AM robot Platform}
\label{sec:MAMbot}
\input{05_MAMbot}

\section{Conclusion and Future Works}
\label{sec:conclusion}
\input{04_conclusion}


\nocite{*}

\bibliographystyle{asmeconf}  
\bibliography{yifeiBib}

\end{document}

%% file: 01_abstract.tex
As the demand for mass customization increases, manufacturing systems must become more flexible and adaptable to produce personalized products efficiently.
Additive manufacturing (AM) enhances production adaptability by enabling on-demand fabrication of customized components directly from digital models, but its flexibility remains constrained by fixed equipment layouts.
Integrating mobile robots addresses this limitation by allowing manufacturing resources to move and adapt to changing production requirements.
Mobile AM Robots (MAMbots) combine AM with mobile robotics to produce and transport components within dynamic manufacturing environments.
However, the dynamic manufacturing environments introduce challenges for MAMbots.
Disturbances such as obstacles and uneven terrain can disrupt navigation stability, which in turn affects printing accuracy and surface quality.
This work proposes a universal mobile printing-and-delivery platform that couples navigation and material deposition, addressing the limitations of earlier frameworks that treated these processes separately.
A real-time control framework is developed to plan and control the robot’s navigation, ensuring safe motion, obstacle avoidance, and path stability while maintaining print quality.
The closed-loop integration of sensing, mobility, and manufacturing provides real-time feedback for motion and process control, enabling MAMbots to make autonomous decisions in dynamic environments.
The framework is validated through simulations and real-world experiments that test its adaptability to trajectory variations and external disturbances.
Coupled navigation and printing together enable MAMbots to plan safe, adaptive trajectories, improving flexibility and adaptability in manufacturing.

%% file: 02_intro.tex
Global manufacturing is shifting from mass production systems to on-demand customized production. 
This change is driven by dynamic consumer demand, unstable supply chains, and the need for sustainability~\cite{kovalenko2024harnessing}. 
Industry 4.0 refers to the integration of digital technologies into manufacturing, including automation, data exchange, and intelligent systems.
Consumers demand products based on their preferences, pushing manufacturers to switch between product variations without costly downtime or redesign workflows.
Smart factories must handle high product variations and personalization while maintaining efficiency and quality~\cite{qin2021self}.
In contrast to traditional production, customization requires fast adaptation to new designs and product configurations~\cite{kovalenko2022cooperative}.
Meeting these demands requires flexible manufacturing processes that could respond quickly to market changes.

Traditional manufacturing systems, such as injection molding or Computer Numerical Control (CNC) machining are based on rigid production lines and slowly follow dynamic market demands.
However, producing new or small-batch products often increases cost and material waste, reducing overall sustainability and efficiency.
These limitations undermine cost-effectiveness and environmental sustainability and do not meet the requirements of Industry 4.0.
Modern factories must balance operational efficiency with flexibility, which calls for advanced manufacturing and control strategies.
Addressing this challenge requires advanced manufacturing solutions.
Advancements in AM has emerged as a key technology for achieving this balance.
Processes such as Fused Deposition Modeling (FDM) and Directed Energy Deposition enable layer-by-layer fabrication directly from digital models~\cite{gibson2021additive}.
These methods reduce tooling needs and allow rapid customization~\cite{liu2022ai}. AM has transformed product design by eliminating the need for dedicated molds and fixtures, reducing downtime and lead time penalties associated with design changes~\cite{bhatt2020expanding}.
For example, to reduce the downtime and improve efficiency, automotive companies are adopting AM to rapidly produce customized grippers and tools directly on the factory shop floor~\cite{zhang2019new}.
AM methods reduce the cost and lead time penalty that traditionally comes with making mass customization.
However, most AM systems remain stationary, confined to fixed build volumes that limit their use in dynamic or distributed manufacturing settings.

The supply chain and production geography are important factors to consider for Industry 4.0.
Mobile robots, including mobile manipulators, add flexibility by transporting materials between fixed manufacturing machines~\cite{chen2018dexterous}.
Mobile robots enable decentralized production, not only in factories, but also closer to the customer~\cite{roca2019technology}.
In factories, these systems are able to handle material transport tasks by dynamically routing parts between workstations via autonomous mobile robots (AMRs)~\cite{unger2018evaluation}.
The routes could be easily reprogrammed through software or redesigned by AI, making it possible to rearrange production workflow~\cite{loganathan2023systematic}. 
This agility enables factories to handle custom orders or adjust production sequences in real-time.
Instead of distributing products from a central factory, companies are able to produce customized items in both regional hubs and during delivery. 
To extend AM’s flexibility, researchers have begun integrating it with mobile robotic platforms, named MAMbots~\cite{li2024mobile}.
MAMbots could perform fabrication directly where it is needed, including within factories or at on-site locations~\cite{weller2015economic}.

MAMbots systems enable on-demand, location-independent production with enhanced adaptability. 
MAMbots navigate in smart factories, collaborate with other robots, and perform tasks, including materials pre-processing, components post-processing, and deposition monitoring~\cite{dorfler2022additive, bhatt2020expanding}. 
Combining AM with mobile robotics increases the design freedom and customization of 3D printing with the flexibility of robotics, enabling adaptable, on-demand production in smart factory environments. 
The combination of AM and autonomous mobility introduces a key shift in Industrial 4.0: from mass production to mass customization, enabled by mobility, digital design, and real-time control systems.

MAMbots aim to overcome these limitations by enabling printers to manufacture while traveling~\cite{li2024mobile}. 
By bringing printers directly to the task location, the system significantly reduces downtime and promises decentralized manufacturing~\cite{kenger2021integrated}.
This fusion of fabrication and mobility also reduces the travel time in production.
Recent research shows that mobile robotic printers could improve overall process efficiency by printing or preparatory steps while moving between workstations~\cite{li2024mobile}. 
MAMbots are capable of creating larger or more complex structures than a stationary 3D printer with enclosure limitation~\cite{dorfler2022additive, li2025model}.
MAMbots have the ability to print beyond the limits of a fixed build box by moving around a workpiece~\cite{lachmayer2022autonomous}.
For instance, researchers have developed mobile robotic extrusion systems that drive autonomously to print large concrete structures directly on construction sites~\cite{keating2017toward}.
Similarly, mobility enables in-situ repair and maintenance by using AM~\cite{dielemans2022mobile}. 
A MAMbot is able to approach damaged machinery or infrastructure and deposit material directly to perform on-site repairs~\cite{weber2022mobile}.
However, maintaining print quality while moving in a real factory environment remains a major challenge~\cite{li2024mobile}.

\begin{figure}
\centering
\smallskip
\smallskip
\captionsetup{belowskip=-1pt}
\includegraphics[width=0.9\columnwidth]{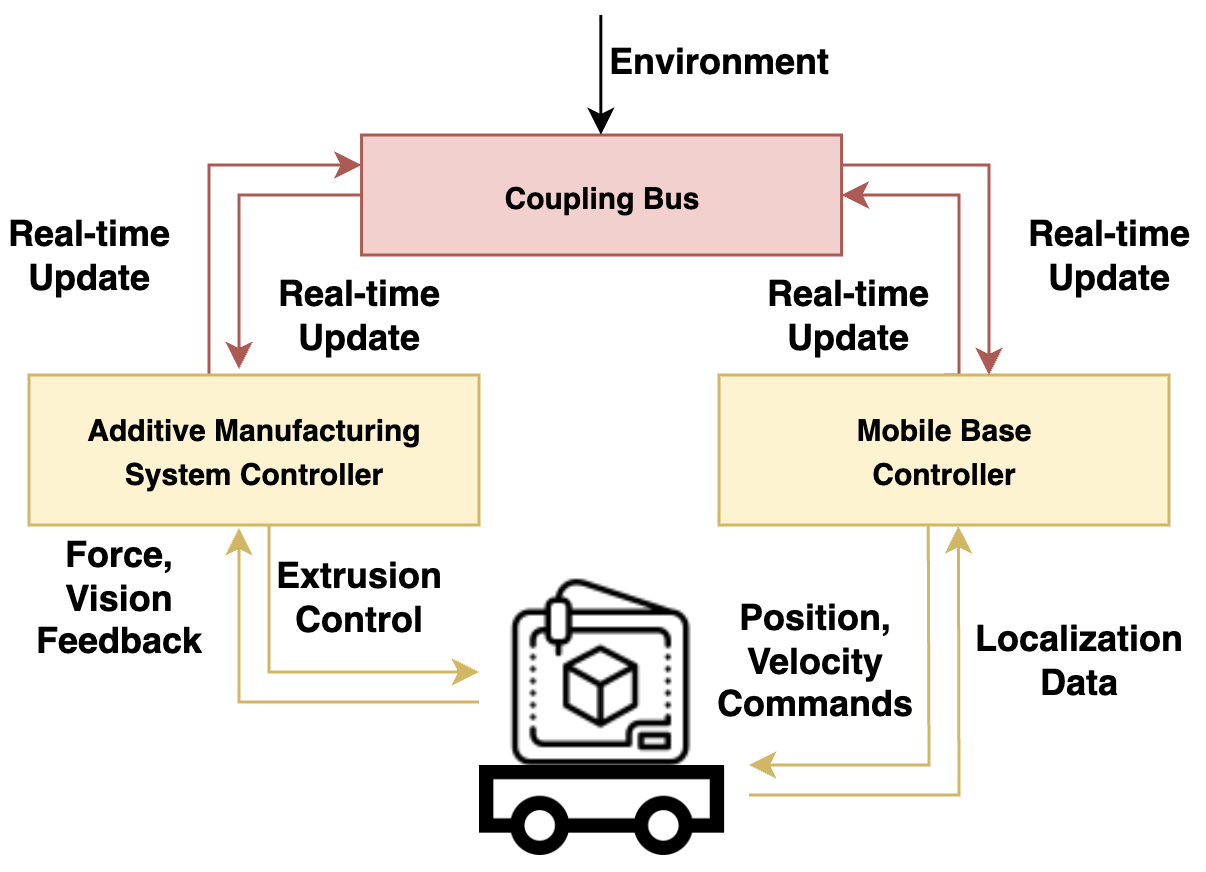}
\caption{Overview of the proposed framework.}
\label{fig:framework}
\end{figure}

Current research focuses on technical challenges such as synchronization between mobility and the manufacturing process, precision during printing and involves complex path planning~\cite{li2024mobile,li2025model}.
However, in real manufacturing environments, robots must operate within constrained spaces, react to unexpected obstacles, and maintain production continuity without reliance on pre-designed behavior.
These environments present compounded dynamic challenges, such as multi-robot and human-robot collaboration.
The coordination of perception, navigation, and material deposition under these dynamic, real-time constraints remains a critical and underexplored capability.
Unlike existing works that treat navigation and fabrication  as separate processes, the proposed framework couples motion control with deposition. Table~\ref{tab:literature} summarizes prior studies and their key limitations.

While mobile printing has been shown in large-scale applications, it is still rarely explored in small or medium-sized smart factories where speed, safety, and flexibility matter.
Unlike traditional stationary AM systems, MAMbots coordinate perception, path planning, and deposition control in real time, requiring strategies that balance safety, motion smoothness, and printing quality.
To address this challenge, both simulation and experimental tests are conducted to evaluate how environmental disturbances affect printing accuracy and the completion time of the task.
The main contributions of this paper are as follows:
\begin{itemize}
    \item A navigation–printing coordination framework that performs safe, detour-aware navigation during the printing process, allowing the system to respond to floor irregularities without halting the task or degrading print quality.
    \item Experimental validation of the proposed approach testing under realistic factory conditions, showing dimensional accuracy improvements of up to 93\% compared to uncoordinated continuous printing.
\end{itemize}

\begin{table*}[t]
\centering
\caption{Summary of prior studies on mobile additive 
manufacturing and related systems}
\label{tab:literature}
\begin{tabular}{p{2.8cm} p{2.2cm} p{2.0cm} p{1.8cm} 
                p{1.8cm} p{4.0cm}}
\hline
\textbf{Study} & \textbf{Platform type} & 
\textbf{Autonomous navigation} & 
\textbf{Navigation--printing coupled} & 
\textbf{Environment} & \textbf{Limitation} \\
\hline
Keating et al.~\cite{keating2017toward} 
& Mobile extrusion robot 
& Pre-planned path 
& No 
& Construction site 
& No dynamic obstacle response; fixed path only \\
\hline
Tiryaki et al.~\cite{tiryaki2019printing} 
& Mobile robot arm 
& Limited 
& Partial 
& Construction site 
& Not validated in factory settings; 
no real-time replanning \\
\hline
Weber et al.~\cite{weber2022mobile} 
& Mobile AM for repair 
& Manual/scripted 
& No 
& Industrial 
& No real-time coordination between 
motion and deposition \\
\hline
Dielemans et al.~\cite{dielemans2022mobile} 
& Mobile clay extrusion 
& Pre-planned 
& No 
& Construction site 
& Static environment only; 
no obstacle avoidance \\
\hline
Lachmayer et al.~\cite{lachmayer2022autonomous} 
& Mobile robot, multi-step AM 
& Autonomous 
& No 
& Construction site 
& No print quality analysis under 
motion disturbances \\
\hline
Bhatt et al.~\cite{bhatt2020expanding} 
& Robotic AM (survey) 
& Varies 
& No 
& Construction site + Industrial 
& Survey only; no dynamic navigation 
or disturbance handling \\
\hline
Li et al.~\cite{li2024mobile} 
& MAMbot (prior work) 
& Autonomous 
& Partial 
& Factory (sim) 
& No terrain disturbance handling; 
no print quality validation \\
\hline
Li et al.~\cite{li2025model} 
& MAMbot with MPC 
& MPC-based 
& Partial 
& Factory (sim) 
& No experimental validation of 
print quality under disturbances \\
\hline
\textbf{This work} 
& \textbf{MAMbot with MPC + detour} 
& \textbf{Real-time, obstacle-aware} 
& \textbf{Yes} 
& \textbf{Factory (sim)} 
& \textbf{Safe, detour-aware navigation with print quality validation} \\
\hline
\end{tabular}
\end{table*}

%% file: 05_MAMbot.tex
\begin{figure*}[tp]
    \centering
    \captionsetup{belowskip=-17pt} 
    \includegraphics[width=1\textwidth]{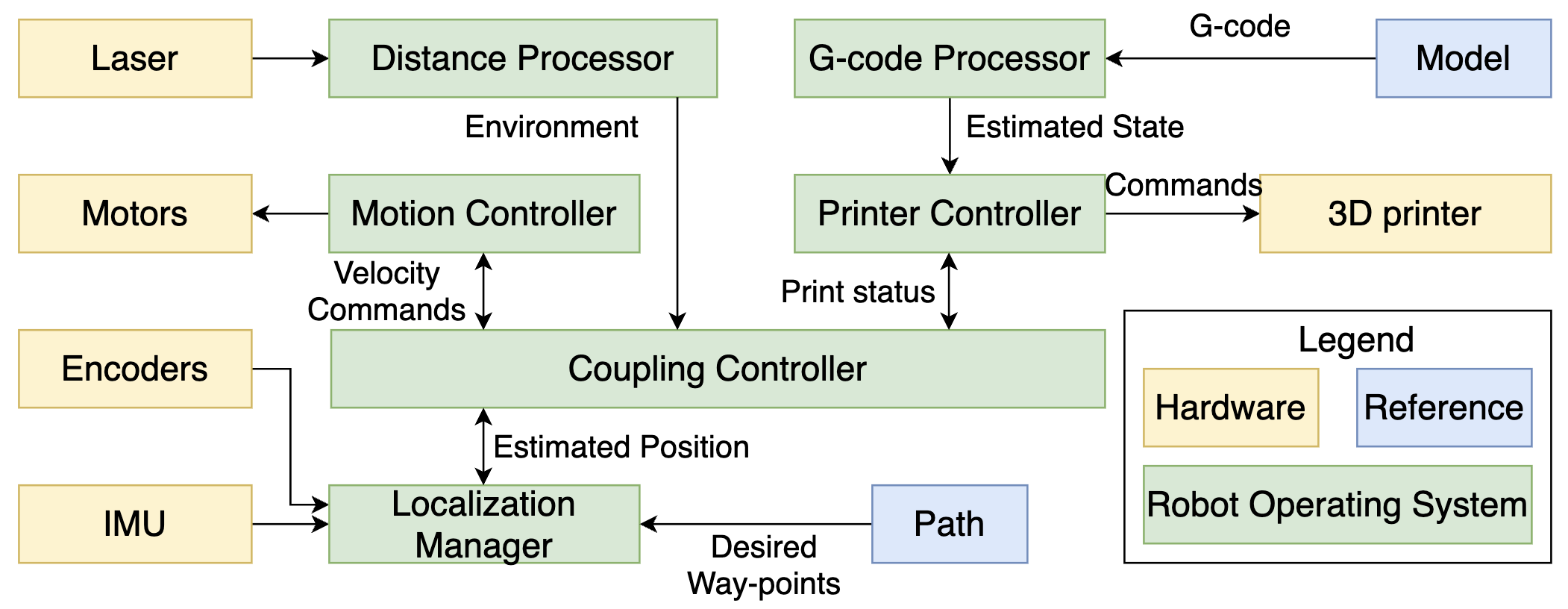}
    \caption{The information flow for the proposed control architecture.}
    \label{fig:flow}
\end{figure*}

This section details the hardware composition and control architecture of the proposed MAMbot system. 
The platform is designed to execute fabrication tasks while navigating real-world factory environments, responding to constraints such as obstacle avoidance, variable terrain, and continuous task execution.

\subsection{Hardware}

This work presents a MAMbot platform designed to support localized, on-demand fabrication in dynamic manufacturing environments. 
The system integrates a differential-drive mobile base, a FDM printer, and a dedicated power source to enable fully untethered operation. 
This configuration allows the robot to navigate freely across the factory floor.
To achieve safe and precise navigation in cluttered industrial spaces, the platform is equipped with LiDAR (Light Detection and Ranging) and inertial measurement units (IMUs) for real-time localization and obstacle detection. 
These sensors support adaptive routing through narrow aisles and congested workspaces where path variability and human-robot interaction.
A FDM printer is mounted on the mobile base and controlled via an onboard processor, enabling coordinated execution of motion and material deposition tasks. 
This design was chosen to ensure tight coupling between navigation and fabrication subsystems.

\subsection{Control Architecture}
To support AM and transportation process coupling, we adopt a hierarchical control architecture composed of two layers: a coupling bus layer, a robot control layer, and a manufacturing execution layer, shown in Figure~\ref{fig:framework}. 
The coupling bus layer acts as the real-time communication backbone, synchronizing data between navigation and printing subsystems and ensuring that changes reflected immediately.
The AM controller regulates key print parameters, including extrusion speed, nozzle motion, and temperature, to maintain consistent print quality.
The moving base controller handles motion planning, localization, and reactive behaviors.
It abstracts sensing and actuation, enabling safe navigation under uncertainty.
Navigation must guarantee not only collision-free travel but also smooth and uninterrupted motion compatible for extrusion-based fabrication.
When the speed is less than the threshold, the effect on print quality is not significant.
However, when the speed exceeds the threshold, the printed parts will have defects or even fail.

Coupled control is essential to preserve print quality, especially during critical operations such as contouring and infill deposition.
During these processes, the platform must adjust its motion to slow down or pause to maintain dimensional accuracy and surface integrity.
A Model Predictive Control (MPC) framework is employed to couple the AM process and the mobile base~\cite{li2024mobile}.
MPC and trajectory smoothing methods are used for short-horizon tracking where print precision must be preserved.
High-level task planning interfaces with production schedules and operator input, enabling dynamic mission allocation.

\subsection{Model Predictive Control}

Navigation between workstations employs MPC for trajectory planning~\cite{li2025model}.
On the manufacturing shop floor, safety-critical areas are commonly divided into zones with different speed limitations.
In addition, the factory environment introduces dynamic challenges.
MAMbots must operate within a range of constraints, including hardware capabilities, physical limitations, obstacle avoidance requirements, and speed restrictions in designated zones.
The MPC framework continuously adjusts its behavior based on the MAMbot's location withing the shop floor map and the quality-speed requirements of the current printing task.
By optimizing trajectory and speed, the system ensures efficient and high-quality manufacturing, even in dynamic factory environments. The proposed controller is formulated as follows:

\begin{subequations}
\label{eq:trajectory1}
\begin{align}
    \operatorname*{min}_{x,u} \quad&
    \operatorname*{cost_{m}}(x,u) + \operatorname*{cost_{p}}(x,u,t_{crit}) 
\\
    \operatorname*{s.t.} \quad
    & \dot{x} = f(x,u), \hspace{4pt} x \notin \mathcal{O}, \hspace{3pt} \forall t \\
    & \dot{x} < \dot{x}_{lim}, \hspace{3pt} x \notin \mathcal{C}, \hspace{3pt}  \hspace{3pt} \forall t \in t_{crit} 
    \vspace{-0.1in}
\end{align}

The cost function is structured as follows:
\begin{align}
cost_{m}(x, u) = 
\sum_{k=0}^{N-1}[(\bm{x}_k - \bm{x}_{\text{ref},k})^T Q ( \bm{x}_k - \bm{x}_{\text{ref},k})]\\+ \bm{u}_k^T R \bm{u}_k \nonumber 
\\cost_{p} (x,u,t_{crit}) = 
\sum_{k=0}^{N-1}\left( \alpha_v \|\ddot{x}(k)\|^2 + \alpha_u \|\dot{u}(k)\|^2 \right)
\end{align}
\end{subequations}
where $x$ are the states of the mobile system (e.g., position, velocity), $u$ is the control input to the system, $N$ is the discrete time horizon of the MPC, $\bm{x}_{\text{ref},k}$ is the reference trajectory, $Q$ and $R$ are positive weighting matrices that prioritize tracking accuracy and control effort, $\alpha_v$ and $\alpha_u$ are parameters for vibration and control smoothness penalties during critical time periods, we penalize acceleration $\ddot{x}(k)$ and its control input $\dot{u}(k)$ rate of change to keep the robot's motion smooth and reduce vibrations that harm print quality.
This coupling is embedded in both the cost functions and the constraints, where the robot’s movement and printing quality are optimized together.

For vehicle dynamics Eq.(1b), $f$ represents a function that captures the dynamics of the mobile system.
In this framework, the system dynamics $f(x,u)$ are typically modeled as a discrete-time system:
$x_{k+1} =A x_k + B u_k$, where $A$ is the state matrix and $B$ is the input matrix from the double integrator vehicle model~\cite{sira2018differentially}. 
For obstacle avoidance Eq.(1b), $k$ is the time step inside the MPC prediction horizon, $\mathcal{O}$ are obstacles in the environment that must be avoided at all times. The $\dot{x}_{lim}$ is allowed maximum velocity during critical printing periods and the $t_{crit}$ is a time interval when the print task is sensitive to motion, which is predefined based on the layer structure of the print task. 
Specifically, the first and last N layers of each print segment are designated as critical intervals, as these layers are most susceptible to adhesion failures caused by motion disturbances. 
For the manufacturing process Eq.(1c), $\mathcal{C}$ denotes certain zones that are strictly constrained during critical times.
$\mathcal{O}$ and $\mathcal{C}$ are encoded as convex polytopics, and enter the constraints to ensure safe motion planning~\cite{jones2010polytopic}.
For the $\operatorname*{cost_{m}}$ Eq.(1d) minimizes the trajectory error based on the system-level planner and existing sensor readings, and $\operatorname*{cost_{p}}$ Eq.(1e) minimizes objectives (e.g., vibrations, turns, etc.) during the critical time window.

In addition to minimizing trajectory error, the cost functions incorporate additional task-specific parameters to reflect printing quality, motion smoothness, and safety constraints.
These parameters regulate print stability and geometric accuracy, particularly during fine-feature fabrication or in environments with dynamic disturbances.
The cost function structure is designed to be weight-tunable, enabling real-time balancing between navigation fidelity and deposition quality.

\subsection{Emergency Detour}
The dynamic nature of the factory environment introduce unpredictable disturbances that can compromise both navigation safety and print quality.
Real-time sensing with LiDAR, IMUs, and cameras supports obstacle avoidance, motion disturbance detection, and precise alignment during fabrication.
These inputs are used for localization, obstacle detection, and precise deposition.
The robot continuously monitors the navigation deviations and the print status.
To integrate reliably into real manufacturing environments, the MAMbot is designed to meet two key operational constraints.
First, it must ensure safe motion in dynamic, human-shared spaces, which demands low-latency decision-making and robust obstacle avoidance.
Second, it must maintain stability on uneven floors and through cluttered or constrained paths.
To meet these demands, the system employs adaptive control and continuous sensor feedback to tolerate motion disturbances while preserving task continuity.

The navigation objective is to maintain nominal forward motion while enabling reactive detours when dynamic obstacles arise. To achieve this, the robot operates under a finite-state motion behavior framework with prioritized response modes based on proximity sensing:
\begin{itemize}
    \item \textbf{Cruise}: Default mode in which no immediate obstacles are detected in the forward field of view. The robot proceeds with nominal forward velocity along the planned trajectory.
    \item \textbf{Slow}: Activated when the frontal clearance distance falls below a predefined safety margin. The robot reduces its speed to a designated low-speed threshold to prepare for potential evasive maneuvers or stopping.
    \item \textbf{Turn}: Triggered when a forward obstacle enters a critical proximity zone. Instead of executing an abrupt stop, the robot performs a rotational maneuver to redirect its heading away from the hazard.
    \item \textbf{Side}: Engaged when the lateral clearance (either left or right) is below the side safety threshold. The robot issues a lateral displacement command to slide away from the encroaching object while maintaining its original heading determined by the frontal logic.
\end{itemize}

The priority hierarchy governing mode selection is as follows: \textbf{Cruise} - \textbf{Slow} - \textbf{Turn} - \textbf{Side}. This ordering ensures that imminent frontal collisions are treated with the highest urgency, while lateral intrusions trigger smooth corrective offsets without disrupting the main navigational intent.

\subsection{Information Flow}

Figure~\ref{fig:flow} illustrates the information flow for the proposed control architecture, showing how sensor data, motion commands, and print instructions are coordinated in real time. Custom Robot Operating System (ROS)~\cite{quigley2009ros} scripts were developed to implement this framework.
A Distance Processor node handles environmental perception, converting raw LiDAR measurements into structured obstacle information for real-time collision avoidance.
The Motion Controller receives velocity or trajectory commands from the coupling layer and translates them into motor control signals, enforcing dynamic constraints based on proximity feedback.
A G-code Processor parses AM instructions and relays them to the Printer Controller, which interfaces directly with the commercial 3D printer. This controller enabling synchronized extrusion control without firmware modifications.
The Coupling Controller combines state information from both motion and printing subsystems to compute joint control strategies.
The Localization Manager integrates odometry, IMUs, and LiDAR data to provide reliable global and local pose estimates in dynamic environments.

\begin{figure*}[t]
\smallskip
\smallskip
    \centering
    \captionsetup{belowskip=-17pt} 
    \includegraphics[width=1\textwidth]{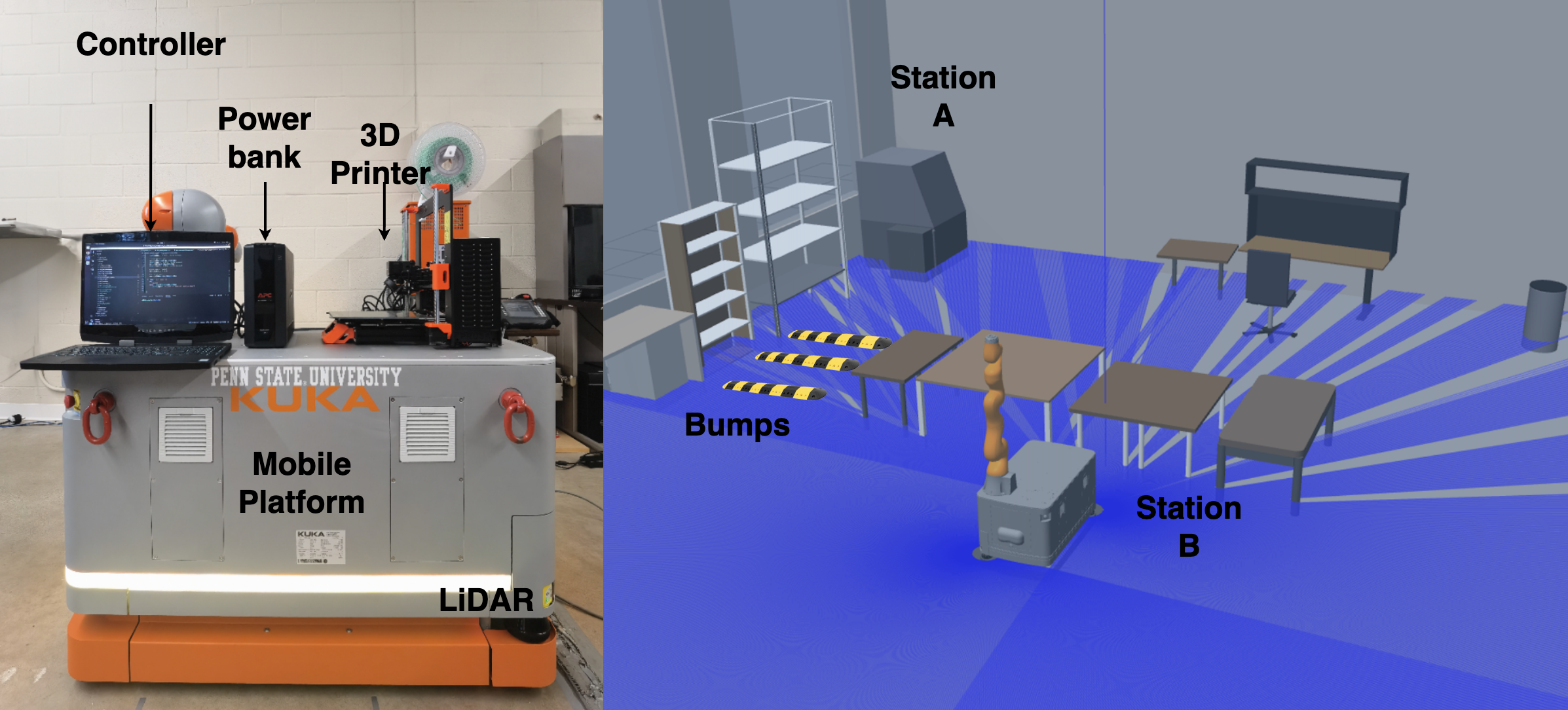}
    \caption{Left: The MAMbot platform consisting of a mobile robot, a commercial 3D printer and a power bank. Right: MAMbots in Manufacturing Environment Simulation}
    \label{fig:platform}
\end{figure*}

\section{Case Studies}

Th case studies aim to evaluate the feasibility and effectiveness of a navigation-integrated MAMbot system deployed in a representative factory environment.
Much of the prior research on mobile 3D printing focused on static motion paths or open spaces.
This work demonstrates a MAMbot performing fabrication tasks while autonomously navigating through constrained industrial layouts with moving obstacles and narrow aisles.
The experiment validates that the robot can maintain continuous task execution while encountering dynamic disturbances during navigation.

\subsection{Feasibility Evaluation with Simulation Platform}

A simulation of the mobile platform was implemented in Gazebo to verify interactions among the mobile robot, sensors, and the manufacturing environment prior to hardware implementation, as shown on the right side of Figure~\ref{fig:platform}.
We tested a KUKA KMR iiwa mobile robot ~\cite{KUKAWeb}  in a simulated manufacturing warehouse environment.
This robot was also used in the hardware implementation.
The experiment was conducted in a 6 × 4 m workspace that included furniture models such as chairs, shelves, power band stands and desks.
The simulated environment contained two workstations connected by a 1m-wide corridor and three small bumps, each with a height of approximately 10 mm placed along the travel path.

The simulations results was used to validate the proposed approach for the mobile system. Specifically, we used it to test whether the robot can navigate the workspace safely and reliably.
The robot was able to move between the two stations, pass through the corridor, and drive over the three bumps without collisions.
These simulations confirm that the controller can support stable and safe motion prior to hardware deployment.
A fixture-printing task was assigned to mimic the real hardware tests, but the simulation only evaluated navigation performance, speed-control logic, and task sequencing.
We then moved to hardware experiments to evaluate actual print quality based on our proposed framework.

\subsection{Hardware Experimental Platform}

We integrated our proposed framework with a KUKA KMR iiwa mobile robot~\cite{KUKAWeb} operating under the Sunrise.OS framework. ROS provided advanced motion planning and real-time communication with the mobile base~\cite{fakhruldeen2021development}. 
The communication architecture relied on ROS and Octoprint Firmware~\cite{rankin2015hack}.
All printed samples shown in this paper come from hardware experiments conducted on the physical KMR iiwa platform with a real Prusa printer.
The fabrication task involved printing a small support bracket (20 × 20 × 12.5 mm). The dimensions of each printed sample were measured using a digital caliper with a resolution of 0.01 mm. Each dimension (X, Y, and Z) was measured three times at the same location, and the average value was recorded to minimize measurement uncertainty.

Environmental perception relies on a pair of wide-angle LiDAR units mounted on the base, providing near-full-coverage obstacle detection.
A commercial Prusa 3D printer, mounted on the KMR platform, served as the additive manufacturing subsystem and performs rapid FDM production. 
To support untethered operation, a power bank supplies dedicated power to the printer, enabling fully mobile printing capabilities.
The assembled platform and its components are illustrated on the left of Figure~\ref{fig:platform}.

\begin{figure}
\centering
\smallskip
\smallskip
\captionsetup{belowskip=-1pt}
\includegraphics[width=1\columnwidth]{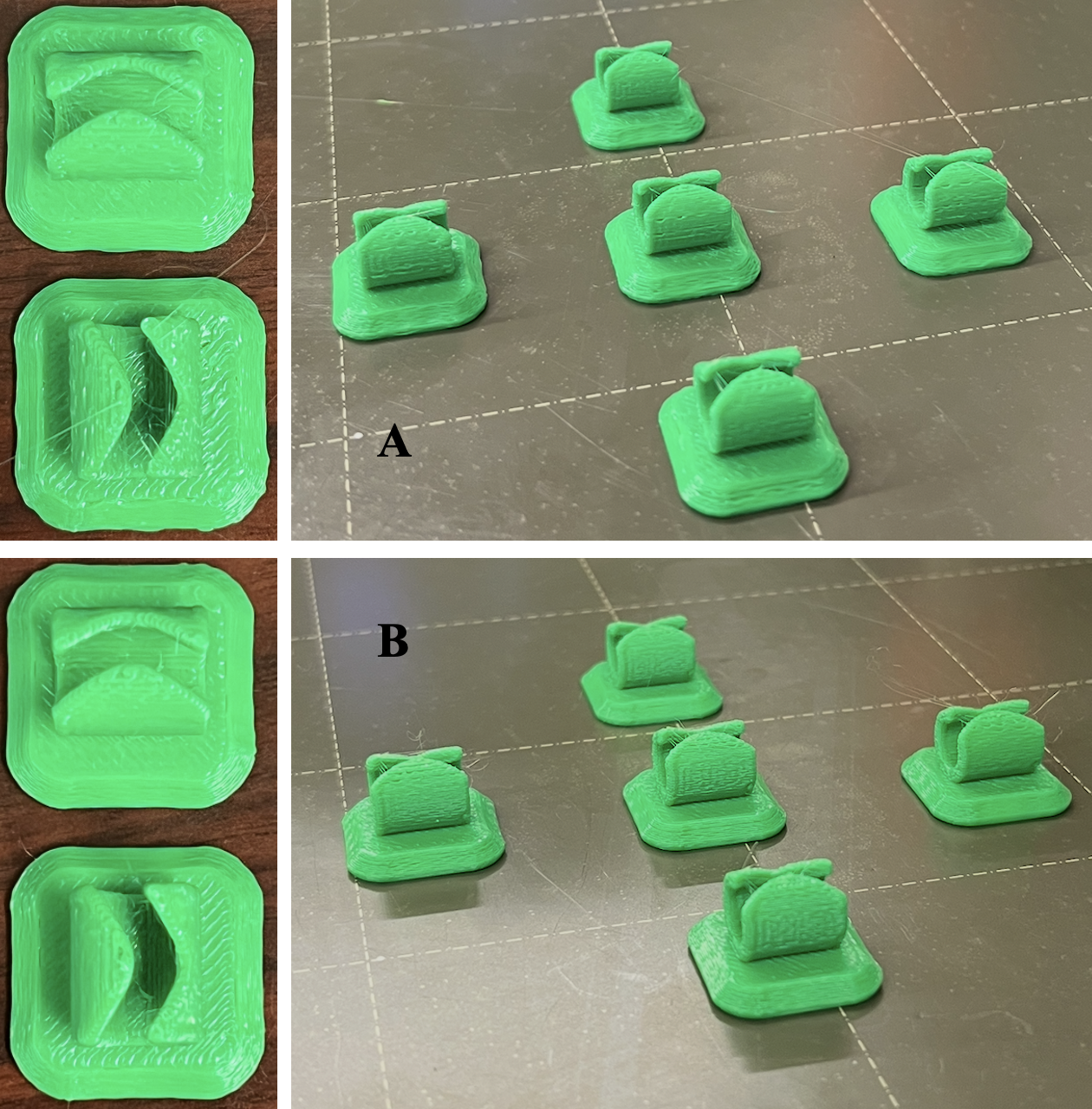}
\caption{Printed samples from Case A and Case B. 
    Top-view surface photos of representative 
    samples from Case A (top, left) and Case B 
    (bottom, left). Samples on the 
    build plate prior to removal, with Case A (top, right) 
    exhibiting visible stringing and layer misalignment, 
    and Case B (bottom, right) demonstrating a cleaner surface 
    finish and improved dimensional consistency.}
\label{fig:result}
\end{figure}

\subsection{Hardware Experimental Procedure}

Two hardware experiments were conducted to evaluate the impact of mobility disturbances on print quality under identical navigation conditions.
In the first trial (Case A), the robot performed continuous printing while traversing three surface bumps, with the Coupling Controller maintaining synchronized base velocity and extrusion flow.
In contrast, the second trial (Case B) implemented a pause-and-resume strategy: the robot halted extrusion before each bump, crossed the obstacle under motion control alone, and resumed printing only after re-establishing stable contact.
This allowed the system to decouple deposition from mechanical disturbances.
Throughout both trials, the printer temperature, the extrusion rate, and the nozzle height were held constant from the G-code. 
The mobile base maintained an average velocity of 0.12 m/s with a speed reduction to 0.06 m/s near the obstacle proximity.

\begin{table}[ht]
\centering
\caption{Raw dimensional measurements of printed parts}
\label{tab:raw_dimensions}
\begin{tabular}{lccc}
\toprule
\textbf{Sample} & \textbf{X (mm)} & \textbf{Y (mm)} & \textbf{Z (mm)} \\
\midrule
Case A1 & 20.77 & 20.92 & 12.60 \\
Case A2 & 21.17 & 20.62 & 12.38 \\
Case A3 & 20.48 & 20.62 & 12.57 \\
Case A4 & 20.61 & 21.13 & 12.72 \\
\midrule
Case B1 & 20.06 & 20.06 & 12.43 \\
Case B2 & 20.10 & 20.08 & 12.35 \\
Case B3 & 20.07 & 20.06 & 12.54 \\
Case B4 & 20.10 & 20.02 & 12.53 \\
\bottomrule
\end{tabular}
\end{table}

\subsection{Results and Discussion}

The printed samples, shown on the build plate prior to 
removal and as surface scans in Figure~\ref{fig:result}, 
Case A exhibited severe layer misalignment and surface roughness caused by transient vibrations transmitted from the bumps to the extruder.
The deposited material over-extruded during vertical shocks and contour distortion.
These defects confirm that continuous printing over uneven terrain induces defects.
Reduction in speed alone cannot fully counteract continuous uneven motion.
By comparison, Case B achieved a clean surface finish, accurate feature geometry, and consistent layer adhesion.
By pausing printing before bump traversal, the system effectively decoupled navigation disturbances from the deposition process.
The pause-and-resume strategy thus prevented cumulative errors while maintaining task continuity and spatial accuracy.

Based on the values quantified from four separate printing  tasks, the resulting dimensional errors are shown in Table~\ref{tab:raw_dimensions} and the comparison of 
averages is shown in Table~\ref{tab:dimensional_accuracy}. 
Case A exhibited average dimensions of approximately  20.76 × 20.82 × 12.57 mm, while Case B achieved  20.08 × 20.06 × 12.46 mm. Relative to the nominal design  size of 20 × 20 × 12.5 mm, Case A showed deviations of  $+$0.76 mm, $+$0.82 mm, and $+$0.07 mm in the X, Y, and Z  directions, corresponding to dimensional error rates of  3.8\%, 4.1\%, and 0.56\%, respectively. Case B reduced these  deviations to $+$0.08 mm, $+$0.06 mm, and $-$0.04 mm, with  error rates of 0.4\%, 0.3\%, and 0.32\%, all within the  ±0.1 mm tolerance threshold. The improvement percentage  is calculated as the reduction in absolute dimensional  error rate relative to Case A, representing improvements  of approximately 89\%, 93\%, and 43\% in the X, Y, and  Z directions, respectively.

\begin{table}[ht]
\centering
\caption{Comparison of dimensional accuracy between Stationary, Case A and Case B}
\label{tab:dimensional_accuracy}
\begin{tabular}{lccc}
\toprule
\textbf{Case} & \textbf{X (mm)} & \textbf{Y (mm)} & \textbf{Z (mm)} \\
\midrule
Design Target     & 20.00 & 20.00 & 12.50 \\
Case A (Avg.)     & 20.76 & 20.82 & 12.57 \\
Case B (Avg.)     & 20.08 & 20.06 & 12.46 \\
Deviation A       & +0.76 & +0.82 & +0.07 \\
Deviation B       & +0.08 & +0.06 & -0.04 \\
\bottomrule
\end{tabular}
\end{table}

These results demonstrate that navigation-induced disturbances, such as bump traversal, can significantly degrade layer alignment and surface quality if not properly coordinated with the printing process.

%% file: 04_conclusion.tex
Manufacturing flexibility influences the ability of a manufacture to respond to market demands, reduce production bottlenecks, and lower inventory costs.
MAMbots represent a convergence of digital fabrication and autonomous systems, allowing the benefits of AM to be delivered wherever and whenever required.
This approach connects key advantages, such as customization, lead time reduction, and production flexibility, to the operational role of mobile AM systems.

This study demonstrated a navigation–printing coordination framework that allowed a MAMbot to perform AM tasks while navigating through a factory environment.
The experiments showed that continuous printing over uneven terrain caused significant dimensional errors.
When no other path reaches the destination, pausing at the appropriate moment and restarting the AM process can improve the printing quality.
The results confirmed that synchronized motion and process control enabled the robot to adjust its behavior in real time.

The development of MAMbots requires research work in a number of areas. 
To effectively use MAMbots, the shop floor needs to be modular and support autonomous navigation. 
Advanced control algorithms must be developed to ensure the reliable operation of MAMbots in dynamic environments.
Integration of MAMbots into existing industrial manufacturing networks also requires standardization in navigation, communication, and scheduling.
For outdoor field deployments, research is needed to improve the flexibility and capability of MAMbots and ensure their operation in unstructured environments.